\documentclass{article}
\usepackage{style/iclr2027_conference}

\usepackage{amsmath,amsfonts,bm}

\def\eqref#1{equation~\ref{#1}}

\def\1{\bm{1}}

\DeclareMathAlphabet{\mathsfit}{\encodingdefault}{\sfdefault}{m}{sl}
\SetMathAlphabet{\mathsfit}{bold}{\encodingdefault}{\sfdefault}{bx}{n}

\newcommand{\E}{\mathbb{E}}

\newcommand{\R}{\mathbb{R}}

\usepackage{times}
\usepackage[T1]{fontenc}
\usepackage[utf8]{inputenc}
\usepackage{microtype}
\usepackage{inconsolata}
\usepackage{graphicx}
\usepackage{booktabs}
\usepackage{tabularx,colortbl}
\usepackage{placeins}
\usepackage{needspace}
\usepackage{float}
\usepackage{amsmath,amssymb}
\usepackage{xcolor}
\usepackage{hyperref}
\usepackage{url}
\hypersetup{hidelinks}

\title{Controlled Acquisition and Abstention in Three-Channel Score Conflicts}

\iclrfinalcopy
\author{Mengzhe Geng\\
National Research Council Canada\\
\texttt{Mengzhe.Geng@nrc-cnrc.gc.ca}}

\newcommand{\ours}{\textsc{Threshold}}
\newcommand{\method}[1]{\textsc{#1}}
\newcolumntype{Y}{>{\centering\arraybackslash}X}
\newcommand{\meanstd}[2]{\shortstack{$#1$\\$(#2)$}}

\begin{document}
\maketitle
\fancyhf{}
\cfoot{\thepage}
\renewcommand{\headrulewidth}{0pt}

\begin{abstract}
When audio, video, and text disagree, accuracy alone does not show whether to acquire another source or abstain. We study these choices in a controlled three-score benchmark: a policy observes two signed scores, may request the third at a cost, and can abstain. The primary reward is mechanism-specific: abstention is correct only for one designated ambiguity mechanism and is penalized under mixed corruption. Matched controls show that a threshold policy matches always-request decisions with fewer requests; its advantage over always-answer fusion depends on the reward assigned to that ambiguity.

On a partially held-out synthetic split, the threshold policy reaches 0.789 +/- 0.006 targeted decision accuracy and 0.481 +/- 0.014 utility across 83 seeds. A three-score majority reference reaches 0.626 +/- 0.008 and 0.252 +/- 0.016, but uses more information. In a matched-budget test, a train-only value selector improves utility over no-query and matched-random policies at 10\% and 25\% budgets, while pair uncertainty has higher utility at every budget. At 50\% and 63.7\% budgets, the selector lowers utility despite slightly higher non-ambiguous accuracy. If all abstentions are scored incorrect, majority outranks the threshold policy in utility. At a central temporal setting, full-trace controls match the neural models while position perturbations separate them.

On held-out-actor emotion clips, eight-frame fusion has opposite-signed accuracy differences for two encoder pairs, with both actor intervals containing zero; matched-request routing gains are small and uncertain. These results separate full-modality accuracy from pre-request selection value and show that selection value depends on budget and the observed-pair ranking.
\end{abstract}

\section{Introduction}
\label{sec:intro}
Audio, video, and text provide complementary evidence, but their combination becomes difficult when one source is corrupted or when the available sources support incompatible explanations. A fusion model that always emits a label hides two operational decisions. It does not say whether a designated missing source is worth acquiring, and it does not say when the observations fail to identify a safe answer. These omissions matter because acquisition, incorrect answers, delays, and abstention can carry different application-dependent costs.

\Needspace{15\baselineskip}
Recent multimodal conflict benchmarks make contradiction detection measurable through controlled perturbations and hierarchical tasks \citep{crosscheckbench2026}. Work on modality interference and credibility-aware fusion shows that source reliability varies across examples \citep{interference2025,credibility2026}. Omnimodal benchmarks also expose counterfactual conflicts across several channels \citep{c3po2026}. Yet a high aggregate accuracy does not specify the value of a missing observation or whether a conflict is identifiable from observations alone. We ask when a third source is worth acquiring, when a conflict should trigger abstention, and whether its value can be predicted before a request.

We study three signed scores representing audio, video, and text. A latent binary event generates the scores before corruption modifies one or more sources. A policy observes two scores, requests the third on disagreement or weakness, and answers when two strong scores agree; otherwise it abstains. Policies never receive mechanism labels. Query and non-ambiguous abstention costs make these decisions measurable. Full-modality value denotes the held-out accuracy change with all modalities observed; pre-request value denotes the gain from selecting requests using only previously observed information. Training and evaluation use separately generated samples, with partly held-out corruption mechanisms.

\subsection{Contributions}
\begin{figure}[h!]
  \centering
  \includegraphics[width=\linewidth]{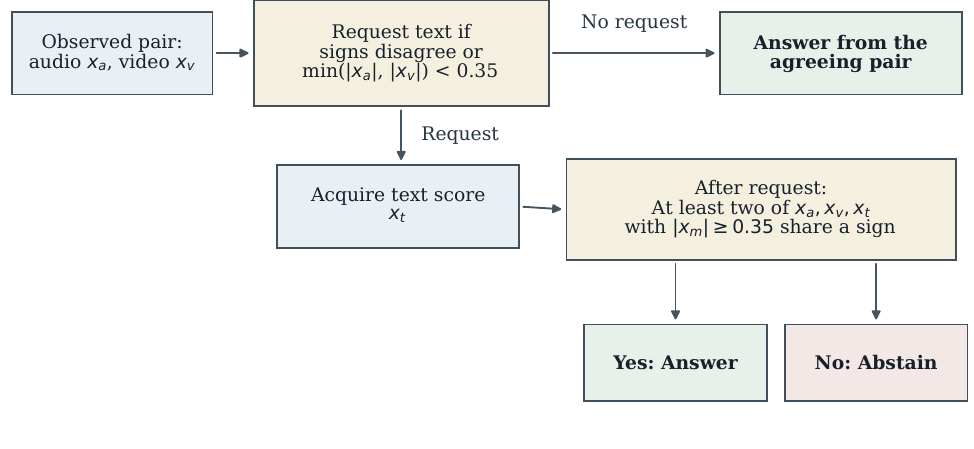}
  \caption{The fixed decision rule. The policy requests text when the audio and video signs disagree or either observed score has magnitude below $0.35$. Without a request, the two strong, agreeing scores determine the answer. After acquisition, the policy answers only when at least two scores have magnitude at least $0.35$ and share a sign; otherwise it abstains. Utility separately charges for a request and for non-ambiguous abstention.}
  \label{fig:problem}
\end{figure}

First, we define a three-score protocol that separates acquisition from the final answer or abstention decision (Figure~\ref{fig:problem}). Matched controls identify when request savings leave final actions unchanged. Second, a common-rule, matched-budget test compares a train-only value selector with pair uncertainty, matched random, and a non-deployable oracle; it shows where a learned ranking helps and where it harms utility. Equal-information temporal controls identify which simulator findings survive matched representations. Third, actor-disjoint media studies separate full-fusion accuracy from selected-request value and show that the results depend on the frozen encoder pair. The evidence supports a controlled decision diagnostic, not a generally superior acquisition policy.

The ConflictBench reproduction code and configurations are maintained in the project repository.\footnote{Project repository: \url{https://github.com/MENGZHEGENG/audio-video-text-conflicts}.}

\section{Related Work}
\label{sec:related}
Multimodal learning has traditionally focused on learning a shared representation or combining modality-specific representations \citep{baltrusaitis2019multimodal}. In conflict settings, this framing is incomplete because the system must reason about whether a source is reliable for the current example. CrossCheck-Bench studies hierarchical image--text contradiction detection in real-world media with controlled perturbations \citep{crosscheckbench2026}. AV-ConfuseBench tests audiovisual confusion created by muting visually present objects \citep{ye2026avconfusebench}.

EmoMM steers attention over supplied audio, video, and text under emotion conflict and missingness \citep{sun2026emomm}.

Our study instead prices a request for a missing score and permits abstention. EmoMM evaluates media models; our primary evidence uses generated scalar scores, with bounded frozen-feature media diagnostics. These tasks and evidence differ.

Modality-interference work diagnoses misleading channels through perturbations and consistency objectives \citep{interference2025}. Credibility-aware fusion estimates instance-level source trust in context \citep{credibility2026}. Source weighting addresses answer formation; acquisition and rejection require additional decisions.

C$^3$PO studies counterfactual conflict across audio, image, video, and text \citep{c3po2026}. We take a narrower experimental route to make action semantics explicit: a three-score generator, explicit mechanisms, and a partially held-out split. This choice sacrifices media realism in exchange for a clear test of whether a policy uses observations alone and whether its errors arise from mandatory answering, unnecessary requests, or non-identifiability.

Selective prediction studies the tradeoff between coverage and predictive risk; SelectiveNet integrates a reject option into a learned model \citep{selectivenet2019}. Active feature acquisition studies purchasing missing information, including generative-surrogate policies \citep{activefeatures2021} and cost-sensitive decision making \citep{costdecision2025}. Probabilistic personalised cascades combine sequential feature acquisition with a rejector that weighs the cost of unobserved features; rejection can trigger another acquisition stage \citep{shpakova2021cascade}. Active Acquisition for Multimodal Temporal Data (A2MT) selects modalities through time under acquisition costs and shows that non-adaptive controls can match or outperform learned agents \citep{kossen2023active}. Acquisition Conditioned Oracle provides a general non-greedy formulation based on expected downstream loss reduction \citep{valancius2024acquisition}. GuessBench studies iterative visual question asking for active reasoning in multimodal large language models \citep{liu2026guessbench}. Our setting fixes two observed scores and one designated unavailable score, and treats a request and terminal abstention as separate decisions. Its distinct evidence is a comparison of action outcomes under specified reward conventions for score conflicts; it does not establish a general acquisition method. Unlike standard selective risk, our synthetic targeted decision accuracy rewards abstention on a designated ambiguity mechanism; the target convention must accompany any comparison.

\section{Problem Formulation}
\label{sec:problem}
Each example has a binary latent event $y\in\{0,1\}$ and an observation vector $x=(x_a,x_v,x_t)\in\R^3$ for audio, video, and text. A clean score is
\begin{equation}
  x_m=(2y-1)s+\epsilon_m, \qquad \epsilon_m\sim\mathcal{N}(0,\sigma^2),
  \label{eq:clean}
\end{equation}
where $s=1$ and $\sigma=0.22$ in the campaign, and write $\mu_y=(2y-1)s$. Labels and mechanisms are sampled uniformly and independently. For uniformly drawn coordinates $j$ and $k\neq j$, invert and swap both set $x_j=-\mu_y+\epsilon_j$; dropout sets $x_j\sim\mathcal{N}(0,\max(1.8\sigma,0.05)^2)$; mixed sets $x_j=-\mu_y+\epsilon_j$ and $x_k\sim\mathcal{N}(0,\max(1.8\sigma,0.05)^2)$; and burst scales the clean vector by 0.7 before setting $x_j=-2.5\mu_y+\epsilon_j$. Ambiguity randomly assigns independent draws from $\mathcal{N}(\mu_y,\sigma^2)$, $\mathcal{N}(-\mu_y,\sigma^2)$, and $\mathcal{N}(0,\max(0.6\sigma,0.04)^2)$ to the three coordinates. The protocol retains seven mechanism labels but only six distinct score-level transformations because invert and swap use the same operation. The training and seen labels are clean, invert, swap, and dropout. The partially held-out split includes independently sampled clean cases and introduces mixed, burst, and ambiguity.

Ambiguity receives an abstention target. This is a mechanism-specific reward convention, not a general definition of non-identifiability. Mixed and ambiguity both contain positive, negative, and zero-centered scores in random positions. Their positive and negative score distributions have equal variance, so swapping the latent label leaves the observation distribution unchanged: $p(x\mid y=0,z)=p(x\mid y=1,z)$ for both mechanisms. With balanced label priors, neither permits observation-only binary accuracy above 50\% in expectation, even with all three sources. Nevertheless, the benchmark rewards rejection only on ambiguity and charges it on mixed. Their zero-centered variances differ, creating a statistical cue for this reward preference. Results therefore measure adaptation to that preference, not universal detection of non-identifiability.

\Needspace{9\baselineskip}
For action $a\in\{0,1,2\}$, where $2$ denotes abstention, correctness is measured against $y$ on non-ambiguous examples and against $2$ on ambiguity examples. The reward also depends on whether the policy requests a third score, indicated by $q\in\{0,1\}$.
We define utility as
\begin{equation}
  \begin{aligned}
  u(a,q)={}&2\mathbb{1}[a=y^\star]-1-\lambda_q q\\
            &-\lambda_a\mathbb{1}[a=2,\;z\neq\text{ambiguity}],
  \end{aligned}
  \label{eq:utility}
\end{equation}
where $y^\star=2$ for ambiguity and $y$ otherwise, $\lambda_q=0.10$, and $\lambda_a=0.20$. Thus a correct action receives $+1$ before costs and an incorrect action receives $-1$; with the campaign costs, per-example utility lies in $[-1.30,1.00]$. We call correctness under this mechanism-conditioned target targeted decision accuracy. We report targeted decision accuracy, coverage, selective accuracy (accuracy conditional on a non-abstaining action), query rate, abstention rate, and mean utility. Means and standard deviations are computed over independent seeds, not over individual examples pooled across seeds.

\section{Protocol and Methods}
\label{sec:methods}
\subsection{Generator and splits}
Each scalar run uses 6,000 training examples and 4,000 examples per evaluation split. Training and seen splits sample clean, invert, swap, and dropout labels with independent seeds. The partially held-out split samples new clean cases, mixed corruption, bursts, and ambiguity; only clean appears in training. Invert and swap implement the same score transformation, giving it twice the sampling and mechanism-macro weight.

\subsection{Fixed and learned baselines}
\method{Majority} predicts the sign supported by at least two scores. \method{Median} predicts the coordinate-median sign. \method{Weighted} predicts the sign of $\sum_m x_m|x_m|$. These policies receive all three scores initially, always answer, and never request a source. The multilayer perceptron (MLP) maps $[x,|x|]$ through $6\!\rightarrow\!32\!\rightarrow\!16\!\rightarrow\!3$ with ReLU hidden units. \method{Gated} uses a tanh-hidden $6\!\rightarrow\!24\!\rightarrow\!3$ softmax gate and a ReLU $2\!\rightarrow\!16\!\rightarrow\!3$ head over weighted mean and score spread. Both learned fusion models also receive all three scores initially and never request a source. These are full-information references, not matched acquisition baselines. Ambiguity is an evaluation-only third class, so neither learned model is fitted as an ambiguity detector.
\method{Learned acquisition} maps the first two scores and their magnitudes through $4\!\rightarrow\!24\!\rightarrow\!12\!\rightarrow\!1$ with ReLU hidden units. Its binary target is whether a third-source request improves training utility after query cost. This target is privileged offline supervision: it is computed from the training target (including the ambiguity target) and the full three-score vector by comparing the no-request pair action with the post-request strong-score action. The scorer itself receives only the initially observed scores and magnitudes at fitting and inference, and its threshold is selected by utility on a separate train-only calibration split. Without a request it answers only when the initial signs agree, otherwise abstaining, and it applies no magnitude threshold. All scalar networks use Adam for 90 epochs, batch size 256, learning rate 0.0015, and gradient-norm clipping at 5; classifiers use cross-entropy and the query scorer uses binary cross-entropy. The scorer is fitted on the four training mechanisms, then selected by utility on 4,000 independent train-only cases over thresholds $\{0.1,0.2,\ldots,0.9\}$. It receives no mechanism, evaluation label, or unavailable text score; post-query decisions use the same strong-score rule as \ours{}.

\Needspace{10\baselineskip}
\subsection{Threshold query-and-abstain policy}
\ours{} starts with the audio and video coordinates, which stand for the currently available pair. It requests the third coordinate when the two signs disagree or when either absolute score is below $\tau=0.35$. This threshold was fixed in the original protocol before held-out evaluation; it was not selected on the partially held-out split or calibrated for optimal utility. Without a request, the policy answers using the first sign. After a request, it answers only when at least two coordinates have magnitude at least $\tau$ and share a sign; otherwise it abstains. This rule uses $x$ only. In particular, it does not inspect $z$, the ambiguity indicator, or the ground-truth label.

\subsection{Matched protocol controls}
All controls start with the same two scores and pay $\lambda_q$ for requesting text. Pair-selective never requests text and abstains whenever the initial pair fails the active rule's agreement/strength test. Query-majority shares the active query schedule but always answers using three-score majority after a request. Always-query-selective always acquires text and then applies the same two-strong-agree rule as \ours{}.

The last control has identical final actions to \ours{} for every input when $\tau>0$. If the active rule requests text, both apply the same decision rule. If it does not, the initial pair already contains two strong agreeing scores, which a single third score cannot overturn. Thus their accuracies and coverage are identical, and their mean utility difference is exactly $\lambda_q(1-\E[q])$. This identity isolates acquisition savings from the advantage of permitting abstention.

\subsection{Matched-budget selector test}
To test whether a request can improve the terminal action, we apply a shared
sum rule before and after acquisition. For an observed set $S$, it answers
with the sign of $\sum_{m\in S}x_m$ when the magnitude is at least
$\tau=0.35$, and abstains otherwise. Unlike \ours{}, the action can change
when the third score is added. At each seed, a standardized ridge model
($\alpha=1$) uses seven features of the initial pair and a training target
equal to the pre-request-cost reward change between applying this rule to the full
triple and applying it to the pair. Labels and the third score define this target only
on the 6,000 training examples; the selector receives the initial pair at
evaluation. We compare its ranking with pair uncertainty (lowest absolute pair
sum first), a seed-specific random ranking, and an oracle that uses evaluation
labels and the unavailable score. The oracle is an upper bound, not a deployable
selector. Every method requests the exact top $k$ of 4,000 held-out examples
at budgets of 10\%, 25\%, 50\%, and 63.7\%, with the same query and
abstention costs. Utility differences use 10,000 paired-seed percentile
resamples with seed 92309.

\section{Experimental Setup}
\label{sec:setup}
The bounded scalar study covers 83 seeds. The temporal factorial has nine cells formed by sequence lengths 32, 64, and 128 crossed with noise levels 0.12, 0.22, and 0.40; each cell uses 23 seeds, for 207 setting-seed runs. Sequence models receive the full $[3,T]$ trace, while scalar controls receive its mean over time. The temporal models use AdamW, width 96, depth 4, batch size 256, and 800 optimizer steps. The temporal evaluation additionally samples contiguous sign inversion, noise occlusion, and channel-local drift. Its seven equally likely mechanisms give an expected ambiguity fraction of one seventh, versus one quarter in the scalar evaluation; their accuracy levels are not directly comparable. Their three-class heads receive no ambiguity examples in training, so the abstention logit is unsupervised; neither temporal model abstained in the evaluation records. No mechanism labels or ambiguity flags are provided as inputs.

The equal-information temporal study fixes $T=64$ and evaluates central $(s,\sigma)=(1.0,0.22)$ and hard $(0.55,0.55)$ settings with five seeds at each setting. It compares the existing temporal fusion and gated networks with a train-only waveform template and a ridge classifier over the same full traces. It also applies a shared time permutation, independent channel permutations, and per-example circular shifts. This study tests whether the initial separation requires the fixed waveform or temporal order.

\Needspace{3\baselineskip}
We use the same utility costs across splits and never tune the active threshold on held-out mechanisms. Acquisition policies start with two coordinates. Full-information scalar references receive three, pay no request cost, and answer every evaluated example.

\Needspace{3\baselineskip}
The seen split tests the training mechanisms. The partially held-out split retains clean cases and adds mixed corruption, high-confidence bursts, and symmetric ambiguity. Comparisons are descriptive when methods use different information or terminal-action rules. We assess answer quality together with coverage, requests, abstention, and utility.

The studies use different analysis units: synthetic seeds, temporal settings, test videos, held-out actors, training-only source-feasibility items, and meeting teams. The six actor allocations overlap and measure split sensitivity; they do not represent independent populations. The AMI evaluation covers 244 donor-matched dialogue-act cases across two teams. Appendix~\ref{app:diagnostics} gives the counts and denominators for the media studies.

\FloatBarrier
\section{Results}
\label{sec:results}
\begin{figure}[!htbp]
  \centering
  \includegraphics[width=0.78\linewidth]{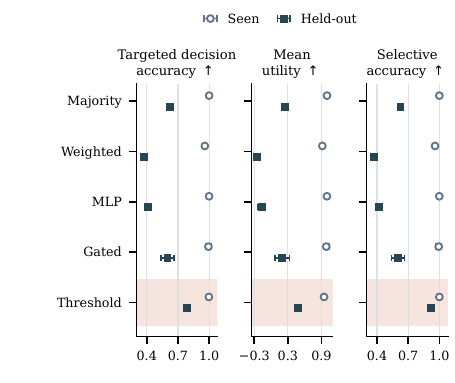}
  \caption{Mean and standard deviation across 83 seeds. The partially held-out split compares full-information fusion references with policies that start from two scores and may request a third. Targeted decision accuracy uses the mechanism-conditioned target defined in the text. Utility includes a 0.10 query cost and a 0.20 cost for non-ambiguous abstention.}
  \label{fig:results}
\end{figure}

Table~\ref{tab:controls} gives the direct acquisition comparison. The threshold rule and always-query-selective make identical final decisions, so their utility difference is exactly the price of requests the threshold avoids. Figure~\ref{fig:results} and Table~\ref{tab:unseen} show the broader synthetic results. On the partially held-out split, the threshold rule reaches $0.789\pm0.006$ targeted decision accuracy and $0.481\pm0.014$ utility. The zero-request majority reference receives all three scores and reaches $0.626\pm0.008$ and $0.252\pm0.016$. This is a full-information reference comparison, not a matched acquisition-policy ranking. Table~\ref{tab:unseen} places the mechanism-conditioned metric beside non-ambiguous binary accuracy, which counts every abstention as an error; on this measure, threshold reaches $0.721\pm0.008$ versus $0.834\pm0.007$ for majority. The learned classifiers receive no ambiguity examples during fitting and serve as additional full-information references.

\subsection{Mechanism-level behavior}
The mechanism breakdown explains the aggregate. On novel ambiguity cases, the active policy abstains at $0.992$, matching the designated rejection target. On burst cases, it retains $0.977$ accuracy and queries on $0.674$ of examples. On mixed cases, mean accuracy is $0.187$, abstention is $0.627$, and utility is $-0.839$. Mixed examples are also label-nonidentifiable, as shown in Section~\ref{sec:problem}; the low utility reflects a reward that still requires a binary answer. It is not evidence that the policy failed to discover an identifiable trusted source.

\begin{table}[!htbp]
\caption{Partially held-out results over 83 seeds. Parentheses give sample standard deviations; rates are seed means with dispersion omitted. Mixed, burst, and ambiguity mechanisms are held out. Targeted accuracy rewards abstention only for designated ambiguity; non-ambiguous binary accuracy excludes ambiguity and counts abstentions as errors. Utility follows Equation~\ref{eq:utility}. Majority, median, weighted, MLP, and gated start with all three scores. Learned acquisition uses pair-only inputs, a full-score training target, and a different no-request action.}
\label{tab:unseen}
\vspace{0.4em}
\centering
\footnotesize
\setlength{\tabcolsep}{2pt}
\begin{tabularx}{\linewidth}{@{}>{\raggedright\arraybackslash}p{1.75cm}YYYYYY@{}}
\toprule
\textbf{Method} & \shortstack{\textbf{Targeted}\\\textbf{acc.} $\uparrow$} & \shortstack{\textbf{Non-ambig.}\\\textbf{binary acc.} $\uparrow$} & \shortstack{\textbf{Utility}\\$\uparrow$} & \textbf{Coverage} & \shortstack{\textbf{Query}\\$\downarrow$} & \textbf{Abstain} \\
\midrule
Majority & \meanstd{0.626}{0.008} & \meanstd{0.834}{0.007} & \meanstd{0.252}{0.016} & $1.000$ & $0.000$ & $0.000$ \\
Median & \meanstd{0.626}{0.008} & \meanstd{0.834}{0.007} & \meanstd{0.252}{0.016} & $1.000$ & $0.000$ & $0.000$ \\
Weighted & \meanstd{0.375}{0.007} & \meanstd{0.499}{0.008} & \meanstd{-0.251}{0.014} & $1.000$ & $0.000$ & $0.000$ \\
MLP & \meanstd{0.415}{0.031} & \meanstd{0.553}{0.040} & \meanstd{-0.170}{0.061} & $1.000$ & $0.000$ & $0.000$ \\
Gated & \meanstd{0.600}{0.063} & \meanstd{0.800}{0.083} & \meanstd{0.200}{0.125} & $1.000$ & $0.000$ & $0.000$ \\
Learned acquisition & \meanstd{0.746}{0.010} & \meanstd{0.756}{0.009} & \meanstd{0.417}{0.019} & $0.709$ & $0.519$ & $0.291$ \\
\rowcolor{green!8}\ours{} & \meanstd{0.789}{0.006} & \meanstd{0.721}{0.008} & \meanstd{0.481}{0.014} & $0.590$ & $0.637$ & $0.410$ \\
\bottomrule
\end{tabularx}
\end{table}
\FloatBarrier

Table~\ref{tab:unseen} makes the action tradeoff explicit. The threshold rule covers 0.590 of cases and abstains on 0.410. Its accuracy conditional on an answer is $0.918\pm0.005$, while majority and median answer every case. A deterministic 10,000-resample paired-seed percentile bootstrap (seed 20270918) gives threshold-minus-majority 95\% intervals of $0.163$ [0.161, 0.165] for targeted decision accuracy and $0.230$ [0.226, 0.233] for utility.

Excluding shared-clean cases, targeted decision accuracy is 0.718 [0.717, 0.720] for the threshold rule, 0.649 [0.647, 0.651] for learned acquisition, and 0.501 [0.499, 0.503] for majority; these are 95\% paired-seed intervals. The corresponding mechanism-label macro values are 0.719, 0.649, and 0.501. This macro average weights mixed, burst, and ambiguity equally.

The reward convention changes the interpretation. In the mixed-reward variant, both mixed and ambiguity cases receive the abstention target, and the abstention cost is waived for both; the threshold rule then has $0.826\pm0.007$ utility versus $0.252\pm0.016$ for majority. In the all-abstentions-error variant, every example targets its binary label and every abstention incurs $\lambda_a$; utility is $-0.064\pm0.019$ versus $0.252\pm0.016$. Thus the utility ordering is a result about the declared target and cost masks, not a general selective-risk advantage.

\subsection{Temporal simulation results}
\label{sec:temporal}
The temporal extension replaces each scalar observation with a zero-mean two-harmonic trace. Its fixed sign-bearing waveform gives sequence models information absent from time means. The 207 setting-seed factorial tests simulator behavior and does not rank general-purpose representations. At length 64 and noise 0.22 on the partially held-out split (including new clean cases), temporal gated reaches $0.787\pm0.006$ targeted decision accuracy and $0.574\pm0.011$ utility; the time-mean MLP reaches 0.434 and time-mean majority reaches 0.417. Across all nine length--noise cells on this split, the cell means range from 0.786 to 0.788 for temporal gated and 0.779 to 0.787 for temporal fusion, versus 0.429 to 0.438 for the time-mean MLP and 0.407 to 0.420 for time-mean majority. These are descriptive ranges across cells, not uncertainty intervals or comparisons with an established chance level. Neither temporal network learned abstention because ambiguity is absent during training.

The equal-information follow-up changes the interpretation: Table~\ref{tab:temporal} shows full-trace template and ridge matching the networks at the central setting. Both controls reach 0.786 targeted decision accuracy and 0.572 utility on the partially held-out split. A shared train--evaluation permutation preserves this result.

\Needspace{6\baselineskip}
Independent channel permutations reduce template and ridge targeted decision accuracy to $0.428\pm0.008$ and $0.428\pm0.015$ while fusion retains $0.768\pm0.009$; per-example shifts give 0.424, 0.430, and 0.781, respectively (Appendix~\ref{app:temporal-permutations}). The hard identity setting barely changes utility, so it supplies no discriminating difficulty gradient. The fixed waveform explains the central and shared-permutation ranking, but the neural advantage under position perturbations remains a simulator-specific finding.

\begin{table}[h!]
\centering
\footnotesize
\caption{Equal-information temporal results at $T=64$, five seeds per setting. All methods receive full traces. Central and shared-permutation settings preserve the positional template; hard identity weakens signal and raises noise. Values are partially held-out targeted decision accuracy or utility, mean (sample standard deviation).}
\label{tab:temporal}
\vspace{0.4em}
\setlength{\tabcolsep}{1.4pt}
\begin{tabularx}{\linewidth}{@{}>{\raggedright\arraybackslash}p{2.65cm}YYYY@{}}
\toprule
\textbf{Method} & \shortstack{\textbf{Central targeted}\\\textbf{decision acc.} $\uparrow$} & \shortstack{\textbf{Central}\\\textbf{util.} $\uparrow$} & \shortstack{\textbf{Shared}\\\textbf{util.} $\uparrow$} & \shortstack{\textbf{Hard}\\\textbf{util.} $\uparrow$} \\
\midrule
Temporal fusion & \meanstd{0.780}{0.011} & \meanstd{0.560}{0.021} & \meanstd{0.571}{0.013} & \meanstd{0.570}{0.016} \\
Temporal gated & \meanstd{0.786}{0.007} & \meanstd{0.571}{0.015} & \meanstd{0.567}{0.018} & \meanstd{0.572}{0.018} \\
Full-trace ridge & \meanstd{0.786}{0.006} & \meanstd{0.572}{0.012} & \meanstd{0.572}{0.012} & \meanstd{0.572}{0.014} \\
Train-only template & \meanstd{0.786}{0.009} & \meanstd{0.572}{0.018} & \meanstd{0.572}{0.018} & \meanstd{0.574}{0.013} \\
\bottomrule
\end{tabularx}

\end{table}

\subsection{Whole-policy learned acquisition diagnostic}
The query scorer receives the initially available audio and video scores and their magnitudes at decision time; changing the unavailable text score leaves these inputs unchanged. Its training target is privileged: it is computed using the full three-score vector and the mechanism-conditioned utility. On seen mechanisms, learned acquisition reaches $0.998\pm0.001$ targeted decision accuracy and $0.954\pm0.004$ utility while querying on $0.425\pm0.034$ of examples. The threshold policy reaches the same targeted decision accuracy, $0.948\pm0.002$ utility, and a query rate of $0.471\pm0.008$. On the partially held-out split, the scorer queries less often ($0.519\pm0.030$ versus $0.637\pm0.007$) but reaches only $0.746\pm0.010$ targeted decision accuracy and $0.417\pm0.019$ utility, compared with $0.789\pm0.006$ and $0.481\pm0.014$ for the threshold policy. Its no-request branch also changes the terminal answer rule, so this whole-policy comparison cannot isolate the value of its query schedule.

\subsection{Cost sensitivity and information budget}
At fixed actions, active utility is
$0.481\pm0.014$ at campaign costs $(\lambda_q,\lambda_a)=(0.10,0.20)$ and
$0.192\pm0.018$ at $(0.40,0.80)$, below majority at $0.252\pm0.016$
(Appendix~\ref{app:cost}). The policy is not retuned across costs, and the
sweep does not measure request latency or media preprocessing.

\Needspace{5\baselineskip}
\subsection{Matched controls and threshold sensitivity}
Across five seeds (4,000 evaluation examples each), always-query-selective
makes the same decisions as the active rule. Their utility gap, 0.03656,
equals avoided-query cost $0.10(1-0.6344)$; query-majority has the same
request rate but lower utility (Appendix~\ref{app:controls}, Table~\ref{tab:controls}).

Excluding shared-clean
cases, active targeted accuracy is $0.720\pm0.006$, utility is
$0.313\pm0.013$, and coverage is $0.461$. A $3\times3$ noise--threshold grid
shows that utility changes sign as the threshold rises (Appendix~\ref{app:cost}).

\subsection{Matched-budget selector value}
Under one terminal rule, train-only value selection exceeds no-query utility
by $0.0138$ [0.0129, 0.0148] at 10\% and $0.0041$ [0.0022, 0.0060] at 25\%;
its differences from matched random are $+0.0249$ and $+0.0303$ (paired intervals in
Table~\ref{tab:matched-selector}). Pair uncertainty has higher utility at every
budget. At 50\% and 63.7\%, learned versus pair-uncertainty non-ambiguous
binary accuracy is 0.4922 versus 0.4827 and 0.4443 versus 0.4296. Non-ambiguous
binary accuracy excludes ambiguity and counts abstentions as errors; utility includes request
and abstention costs; learned utility is lower at both budgets. Among requested examples, mean reward change from
the third score before query cost is $-0.0591$ versus $+0.1756$ at 50\% and
$-0.1343$ versus $+0.0235$ at 63.7\% (learned versus pair uncertainty).
These are within-seed request means across 83 seeds, not independent population
units; high-budget learned requests reduce reward. Learned-minus-no-query
utility differences are $-0.0795$ [$-0.0825$, $-0.0766$] and $-0.1493$
[$-0.1523$, $-0.1462$]. The oracle uses held-out labels and the unavailable
score (Appendix~\ref{app:selector}, Table~\ref{tab:matched-selector}).

\subsection{Oracle value versus pre-query cue}
Across 24 seeds of a two-candidate simulator, oracle gain is 0.10 at a 0.50
request rate, while cue gain over matched random ranges from 0.000019 with no
cue to 0.099953 with a perfect cue (Appendix~\ref{app:identifiability}).
Oracle headroom alone does not establish feasible pre-request value.

On CMU-MOSEI descriptors \citep{mosei2018}, calibration reaches 65.17\% accuracy versus 65.29\% for same-request majority. Task-fit projections of released descriptors meet the head-adequacy criteria, but the router's error-reduction difference from the calibration-selected matched active baseline is 0.000499 (95\% video-group interval [$-0.000439$, $0.001908$]) at a 0.50 request budget, below the prespecified 0.005 margin; its video-macro difference is $-0.000156$. An online threshold reduces error from no request by 0.025227 at a 0.490754 request rate. These analyses use descriptors, not raw media (Appendices~\ref{sec:mosei} and~\ref{app:mosei-actions}).

The Perception Test training pool does not meet the source-grounding threshold and contains 12 unseen components, below the 15-component minimum; it therefore does not support a conflict benchmark or held-out evaluation (Appendix~\ref{sec:perception}).

\subsection{Natural-media diagnostics}
On 1,244 actor-held-out CREMA-D clips \citep{cao2014cremad}, eight-frame fusion minus audio accuracy is $+5.87\%$ (95\% actor interval [$-0.56\%$, $+11.77\%$]) for wav2vec 2.0 \citep{baevski2020wav2vec2} with Vision Transformer \citep{dosovitskiy2021vit}, and $-2.49\%$ ([$-8.40\%$, $+2.81\%$]) for Hidden-Unit BERT \citep{hsu2021hubert} with Data-efficient Image Transformer \citep{touvron2021deit}. Router minus matched-random accuracy is $+0.04\%$ at 99.36\% requests and $+0.42\%$ at 71.70\%; both intervals span zero (Appendix~\ref{sec:cremad}, Table~\ref{tab:cremad-main}). In the two-team AMI evaluation, audio--video--text accuracy is 35.77\% versus 44.60\% for audio--video; text routing minus static audio--video accuracy is $-3.69\%$ at 15.57\% requests (Appendix~\ref{app:ami}). Neither study establishes a reliable selective-acquisition gain.

\enlargethispage{1\baselineskip}
\section{Conclusion and Limitations}
\label{sec:limits}
The threshold rule saves requests under its synthetic reward. Learned
selection helps only at low budgets and loses to pair uncertainty in utility.
Temporal findings depend on the waveform; natural-media acquisition gains,
latency, privacy, and perceptual benefit remain unconfirmed. Human evaluation
is future work (Appendix~\ref{app:limitations}). 

\clearpage
\section*{AI Use Statement}
We used AI tools to help polish the writing and find related work. Details are
described in the paper.

\section*{Reproducibility Statement}
Sections~\ref{sec:methods} and \ref{sec:setup} and Appendices~A--E provide the
evaluation details. The project repository contains the ConflictBench implementation and setup
instructions; project-owned code is released under
the MIT License. Dataset and dependency terms remain with their respective
providers.

\clearpage

\bibliographystyle{style/iclr2027_conference}
\bibliography{refs}

\appendix
\section{Extended Transfer and Source Diagnostics}
\label{app:diagnostics}
This appendix follows two questions: what changes when another source is
available, and whether its benefit can be anticipated before requesting it.
CMU-MOSEI tests the scalar rule on released descriptors and screens request
policies with separate fitting groups. A controlled simulator distinguishes
hidden oracle value from an observable pre-request cue. The training-only Perception Test study does not support a source-grounded conflict benchmark. CREMA-D compares audiovisual fusion and matched-request
routing across frozen encoder pairs, while the AMI pilot evaluates three modalities across two held-out
meeting teams. The later sections examine
synthetic action controls, matched request budgets, and cost sensitivity.
None of the media studies establishes a reliable gain from selectively
requesting a source.

The study counts refer to different objects. The scalar study uses 83 seeds and 4,000 examples in each evaluation split per seed. The temporal
factorial uses 23 seeds at each of nine length--noise settings, for 207 setting-seed runs; its five-seed full-trace test is a separate follow-up. Each of four
controlled pre-query simulation families has 24 seeds and six cue-strength
settings, with 400,000 generated evaluation examples per seed at each
setting. CMU-MOSEI scores utterances but resamples 676
test videos for the reported transfer interval. The descriptor-routing
screens instead reuse the same 1,871 validation utterances from 300 videos
across fitting seeds and protocols; they are not independent data
replications. The Perception Test source study uses 100 training-only targets, 200 target--donor pairs, and 2,800 generated answers; these analysis units are distinct from independent examples. CREMA-D scores 1,244 clips and resamples 16
evaluation actors, with six overlapping role allocations used only to probe
sensitivity to the allocation. AMI evaluates 244 donor-matchable dialogue-act cases across two meeting teams and reports their unweighted mean. None of these media units is exchangeable with a synthetic seed or a temporal setting.

\subsection{CMU-MOSEI Descriptor Transfer}
\label{sec:mosei}
We test the unchanged rule on released CMU-MOSEI computational-sequence
descriptors \citep{mosei2018}: COVAREP audio, OpenFace2 visual, and Timestamped
Word Vectors text features. For each labeled utterance, we average temporally
overlapping features, standardize with training utterances, and project onto the
normalized positive-minus-negative training class-mean direction. The audio,
visual, and text descriptors have 74, 713, and 300 dimensions. Sentiment below
zero is negative; other values are nonnegative. This is projected-descriptor
sentiment, not raw-media inference or a corruption benchmark.

The official video folds yield 16,327/1,871/4,662 train/validation/test
utterances shared across modalities. Two test videos lack a complete feature intersection and are excluded. Binary targets use the original utterance sentiment labels: values below zero are negative, and values at or above zero are nonnegative. Projections use no validation or test labels. Five seeds use batch size 256, learning rate 0.0015, and 40 epochs. The fixed rule keeps
$\tau=0.35$ without target-domain tuning.

Table~\ref{tab:mosei} shows that the synthetic ranking does not transfer. The
threshold rule obtains 0.624 test accuracy, 0.944 coverage, and 0.417 query
rate; majority reaches 0.653 and the multilayer perceptron (MLP) reaches
$0.773\pm0.002$. A deterministic 10,000-resample bootstrap over 676 test videos
gives 95\% intervals of [0.597, 0.651] for the threshold rule and [0.627,
0.679] for majority; their paired difference is $-0.029$ [$-0.035$, $-0.023$].
All models receive only three projected scores, so this is not a comparison
with current multimodal encoders. MLP deviation is training-seed variability on
fixed folds; fixed policies do not vary across seeds. We add no synthetic
ambiguity target or corruption.

\begin{table}[h!]
\centering\footnotesize
\caption{CMU-MOSEI descriptor accuracy on official folds. Five seeds; only
learned models vary. These are not corruption-family splits.}
\label{tab:mosei}
\begin{tabularx}{\linewidth}{@{}>{\raggedright\arraybackslash}p{2.5cm}YY@{}}
\toprule
\textbf{Method} & \textbf{Validation accuracy $\uparrow$} &
\textbf{Test accuracy $\uparrow$}\\
\midrule
Majority / median & 0.631 & 0.653\\
Weighted & 0.587 & 0.616\\
MLP & \meanstd{0.773}{0.002} & \meanstd{0.773}{0.002}\\
Gated & \meanstd{0.773}{0.002} & \meanstd{0.772}{0.002}\\
\rowcolor{green!8}\ours{} & 0.598 & 0.624\\
\bottomrule
\end{tabularx}
\end{table}
\FloatBarrier

As a calibration check, we select each control's threshold by validation
utility over 11 values from 0.01 to 2.00, breaking ties downward and leaving
test labels unused. Every control selects the grid minimum, so this does not
establish an optimum outside the grid. The calibrated rule reaches 65.17\% test
accuracy, 0.2659 utility, 0.3700 query rate, and 0.9981 coverage. The
same-request majority control reaches 65.29\% and 0.2689; always acquiring text
with the selective rule has identical accuracy but 0.2029 utility. Calibration
explains much of the fixed rule's accuracy deficit, while acquisition savings
remain distinct from final decisions.

\subsection{Projected-Descriptor Action Screen}
\label{app:mosei-actions}
A separate five-seed, cost-sensitive action screen uses utterance-aligned projected features. It divides official training videos into task-fit, router-fit, and calibration groups and does not use the official test fold. The three-score projection itself was fit
with labels from all official training groups before that division. This is
an exploratory downstream screen; its projected features do not constitute an
independent feature-learning test. At a fixed 0.50 acquisition budget, an
incorrect answer costs 1,
abstention costs 0.25, and a request costs 0.05 in normalized loss units.
Table~\ref{tab:mosei-actions} reports cost reduction relative to making no
request, with 95\% intervals formed by resampling validation-video groups.
All five seeds have negative absolute differences for each task head, but each
interval includes zero. The data therefore support neither a positive
acquisition result nor a strictly harmful effect. This screen uses released descriptors, not raw media or a natural conflict label.

\begin{table}[h!]
\caption{Projected CMU-MOSEI direct-action screen at a 0.50 request budget.
Mean normalized cost reduction versus no request over five seeds; paired
video-group intervals use 10,000 bootstrap draws. Positive values favor
acquisition.}
\label{tab:mosei-actions}
\centering\footnotesize
\begin{tabularx}{0.85\linewidth}{@{}>{\raggedright\arraybackslash}p{3.15cm}YY@{}}
\toprule
\textbf{Task head} & \textbf{Cost reduction $\uparrow$} & \textbf{95\% interval}\\
\midrule
Modality-dropout ridge & $-0.0055$ & $[-0.0126,\,0.0017]$\\
Separate-mask ridge & $-0.0041$ & $[-0.0112,\,0.0032]$\\
Shared mask-aware ridge & $-0.0055$ & $[-0.0128,\,0.0017]$\\
\bottomrule
\end{tabularx}
\end{table}

A second five-seed screen evaluates a value router that uses only available
features at decision time and a calibration-fixed online threshold on the same
projected cache. With one observed source and two possible sources to request,
the selected router's error-reduction margin over its selected active baseline
at the 0.50 target budget is 0.000748 (video-group 95\% interval
[$-0.000710$, 0.002339]). With two observed sources and one possible request,
the margin is 0.000784 ([$-0.001203$, 0.003330]). Both miss the preregistered
0.005 margin and have intervals crossing zero. Relative to no request, the
online threshold reduces error by 0.022911 at a 0.487333 realized request
rate with two candidates; with one candidate, its reduction is 0.013932 at
a 0.504935 rate. Neither comparison establishes an advantage over the matched
active baseline. At the same 0.50 budget, a label-informed oracle reduces
error by 0.074755 with two candidates and 0.041511 with one. It uses outcomes
unavailable before the request. This headroom does not yield a supported
allocation gain under this screen.

For the audio-observed, two-candidate decision in this projected-descriptor
screen, the online router requested text 6,768 times and video 42 times
across the five fits at the 0.50 target budget. The realized-value sum for video requests was zero in each fit, compared with a total of 216 for text requests. These decisions share validation videos and are not independent examples; they do not establish that video has no value in general.

The second value study uses pooled, utterance-aligned COVAREP audio (74 dimensions), OpenFace2 video (713), and timestamped word vectors (300).
The released descriptors are unprojected. For each seed, we
split videos first, then standardize and project each source to one score
using a class-mean direction fitted only on task-fit videos. We fit task heads
there, then train the router and select its baseline on separate groups.
Five seeds each cover two observed sources with one possible request and one
observed source with two possible requests. The task heads meet adequacy criteria, while the two-candidate oracle leaves headroom. For the two-candidate decision at an exact
0.50 request budget, the selected router improves error reduction over its
calibration-selected active baseline by 0.000499 (95\% video-group interval
$[-0.000439,\,0.001908]$). This misses the registered 0.005 margin; the
video-macro difference is $-0.000156$. The gain by initial source is 0.001924
for audio, 0.000107 for text, and $-0.000534$ for video. An online threshold
fitted on calibration videos reduces error against no request by 0.025227
(95\% interval $[0.005862,\,0.045641]$) at a realized 0.490754 request rate.
That comparison does not show improvement over the matched active baseline.
At the same exact 0.50 budget, the label-informed oracle's error
reductions are 0.078675 with two candidates and 0.042188 with one;
it cannot serve as a deployable pre-query policy.

\Needspace{4\baselineskip}
A distinct value study splits videos before standardization, projection, and router fitting, avoiding the feature-learning overlap in the preceding screen. Both analyses use released computational descriptors, not raw-media encodings.

\FloatBarrier

\Needspace{8\baselineskip}
\subsection{Controlled Pre-Query Value Simulations}
\label{app:identifiability}
The controlled simulator fixes one helpful and one harmful candidate at values
$+0.20$ and $-0.20$ before costs. Its observable cue identifies the helpful
candidate with probability $(1+\rho)/2$, so the oracle has information that
the policy lacks. Each of four studies uses 24 seeds and 400,000 evaluation
examples per seed. The first uses no acquisition cost and an exact 0.50 query
rate for the cue, random, and oracle policies. Oracle mean gain is 0.10 at
every $\rho$; cue-minus-random mean gain is 0.000019 at $\rho=0$, 0.039848 at
$\rho=0.4$, and 0.099953 at $\rho=1$. The cue advantage over random grows as cue accuracy rises, as expected from the simulator design.

The second study charges both candidates 0.02. At the same exact 0.50 query
rate, cue-minus-random gain is $-0.000067$ at $\rho=0$ and 0.099995 at
$\rho=1$. Its risk-aware abstention rule has no exact-budget constraint; at
$\rho=1$ it queries every example and has mean net gain 0.18. The third
study estimates candidate-specific lower gains from a separate 400,000-example
calibration sample, with candidate costs 0.02 and 0.10. The calibrated policy
queries none at $\rho=0$, one candidate on roughly half the examples at
$\rho=0.4$ (mean query rate 0.499899, net gain 0.030002), and both candidates
at $\rho=0.8$ (query rate 1.000000, net gain 0.099993). Replacing its
calibration sample with 400 examples agrees at the three prespecified cue strengths: query rates 0, 0.499954, and 1.000000 at $\rho=0$, 0.4, and 0.8,
with net gains 0, 0.030001, and 0.099999. Agreement does not extend across
the inspected cue-strength grid. At $\rho=0.2$, the large-calibration policy
requests at rate 0.499850, whereas the 400-example policy makes no requests.
At $\rho=0.6$, their rates are 1.000000 and 0.687523, respectively; nine of
the 24 finite-calibration runs request on every example, while the other 15
request on roughly half. Their mean net gains are 0.060004 and 0.053745,
respectively. The calibration sizes use different seed families, so these
differences are descriptive, not paired estimates of a calibration effect.
These cost-sensitive policies can exceed the 0.50 request rate of their random and cue comparators. The simulator does not evaluate natural-source acquisition or establish an advantage at a fixed request budget.

The contrast is between information that exists after acquisition and a cue
available beforehand. An oracle gain alone cannot justify a request policy:
with an uninformative cue the same-budget policy remains near random, while
the unconstrained calibrated policies can request more often than their
comparators.

\subsection{Raw-Media Source-Feasibility Study}
\label{sec:perception}
We test whether Perception Test training data \citep{patraucean2023perception}
support source-specific questions for a controlled audiovisual extension.
Qwen2.5-Omni-3B revision \texttt{f75b40e3} \citep{xu2025qwenomni,qwenomni3bmodel}
answers the original choices from audio, video, or aligned audiovisual input
using deterministic four-token decoding; video uses 2 frames per second (fps)
and 4--32 frames. The training-only design has 100 targets and 200
target--donor pairs, split by component into 60/20/20 targets for fitting,
calibration, and evaluation. Its decision partition has 40 pairs.
Question-only and same-question shuffled-media controls measure agreement without
aligned input. Generated answers are mapped to the original choices with a fixed grammar.

Before inference, we set feasibility thresholds of 70\% source accuracy, 50\%
pair support, and a 20\% absolute aligned-over-control accuracy gain.
Target/donor accuracy tests unique questions; pair support requires both
questions correct. Gain subtracts the larger question-only or shuffled-media
agreement from aligned accuracy. These thresholds are fixed before inference. Table~\ref{tab:perception-source} shows
30--40\% target accuracy, 40--55\% donor accuracy, 15--25\% pair support, and
an absolute aligned gain of at most 10\%. For 20 pairs, pair-support Wilson 95\%
intervals are \mbox{5.2--36.0\%}, \mbox{8.1--41.6\%}, and \mbox{11.2--46.9\%} at 15\%, 20\%, and
25\%. All nine paired-component bootstrap intervals for gain include zero, with
endpoints from $-20\%$ to $30\%$ in absolute accuracy differences. The
source-grounding criteria are not met, and the component split contains 12
unseen components, below the prespecified minimum of 15. The training pool
therefore does not support a source-grounded conflict benchmark under this
design. No validation/test data, raw routing, or conflict evaluation are
included. This result is not a general Perception Test accuracy estimate.

\begin{table}[h!]
\centering\footnotesize
\caption{Perception Test source-feasibility results (percent; target shows Wilson
95\% CI; donor/pair rates use 20 pairs). A+V is aligned input; T/S/O is
target/same/opposite. Minima: 70 source accuracy, 50 pair support, and
an absolute accuracy gain of 20\%.}
\label{tab:perception-source}
\vspace{0.4em}
\setlength{\tabcolsep}{1.5pt}
\begin{tabularx}{\linewidth}{@{}>{\raggedright\arraybackslash}p{1.0cm}YYYY@{}}
\toprule
\textbf{Input} & \textbf{Target (\%) [CI] $\uparrow$} & \textbf{Donor S/O (\%) $\uparrow$} & \textbf{Pair S/O (\%) $\uparrow$} & \textbf{Gain T/S/O (\%) $\uparrow$} \\
\midrule
Audio & 30 [15, 52] & 50 / 55 & 25 / 15 & 0 / 10 / 5 \\
Video & 40 [22, 61] & 40 / 55 & 20 / 20 & 5 / 0 / 5 \\
A+V & 40 [22, 61] & 40 / 55 & 20 / 20 & 5 / 0 / 5 \\
\bottomrule
\end{tabularx}

\end{table}

\FloatBarrier
\subsection{CREMA-D Natural Audio--Video Evidence}
\label{sec:cremad}
CREMA-D provides paired WAV and FLV media, independent voice and face votes,
and integrated emotion votes under the ODbL-1.0 database and DbCL-1.0 contents
terms specified by the official release \citep{cremadrepo}. We use its 14,884
paired media files. This is an audio--video representation diagnostic within an
audio--video--text protocol paper, not a natural-text evaluation. Before label
inspection, a deterministic assignment allocates 91 actors to 45 task-fit, 15
router-fit, 15 calibration, and 16 evaluation actors. The primary target uses the 7,022
singleton integrated votes. We retain the 420 tied votes for a set-valued
sensitivity analysis. Clip identifiers are pairing keys only, and the fixed
sentence context is not an emotion feature.

We freeze \texttt{facebook/wav2vec2-base-960h} audio features and
\texttt{google/vit-base-patch16-224} visual features, fit standardized logistic
heads on task-fit actors, and mean-pool uniformly sampled visual frames. The
router uses only observed-modality features, is fit on router-fit actors, and
chooses its threshold on calibration actors. Its matched random comparator has
the same evaluation request rate $p$ and expected correctness
$p\,\mathrm{correct}_{A+V}+(1-p)\,\mathrm{correct}_{A}$.
Table~\ref{tab:cremad-temporal} shows that one visual frame is reliably worse
than audio alone, while four and eight frames have positive but imprecise
actor-cluster intervals on the initial role allocation. Across five more
overlapping actor allocations, eight-frame fusion has absolute accuracy gains
of 4.75--8.38\%; these are split-robustness checks, not independent populations.

\begin{table}[H]
\caption{CREMA-D actor-held-out singleton evaluation with frozen wav2vec 2.0
and Vision Transformer (ViT) features. A+V denotes audio--video fusion. Every
condition uses the same 45/15/15/16 actor roles and 1,244 evaluation clips;
only the number of uniformly sampled visual frames changes. The last column
is the absolute accuracy difference (A+V minus audio) computed from unrounded
scores; intervals use 10,000 paired bootstrap resamples of the 16 evaluation
actors.}
\label{tab:cremad-temporal}
\vspace{0.4em}
\centering
\footnotesize
\begin{tabularx}{0.92\linewidth}{@{}>{\raggedright\arraybackslash}p{2.4cm}YYY@{}}
\toprule
\textbf{Visual frames} & \textbf{Video acc. (\%) $\uparrow$} & \textbf{A+V acc. (\%) $\uparrow$} &
\shortstack{\textbf{A+V $-$ audio}\\\textbf{absolute diff. (\%) $\uparrow$}}\\
\midrule
1 & 22.03 & 35.45 & $-9.65$ [$-13.70$, $-5.28$]\\
4 & 37.78 & 48.15 & $+3.05$ [$-2.80$, $+8.21$]\\
8 & 43.09 & 50.96 & $+5.87$ [$-0.56$, $+11.77$]\\
\bottomrule
\end{tabularx}
\end{table}
\FloatBarrier

To probe encoder dependence, we replace wav2vec 2.0 audio with Hidden-Unit
BERT (HuBERT) and Vision Transformer (ViT) video with a Data-efficient Image
Transformer (DeiT) in the eight-frame procedure. The media, actor roles,
labels, prediction heads, and router protocol remain fixed. Audio is decoded
as 16-kHz mono float32 and mean-pooled from the last hidden state; video frames
are sampled at uniform points spanning the full clip interval and processed
by the model image processor; class-token embeddings are mean-pooled. For wav2vec 2.0, we restore the audio positional-convolution weights from the
checkpoint's legacy parameterization before feature extraction.
The HuBERT--DeiT diagnostic gives 56.43\% audio--video accuracy versus
58.92\% audio-only accuracy, an absolute accuracy difference of $-2.49\%$
with a 16-actor bootstrap interval of [$-8.40\%$, $+2.81\%$]. Across the
same five deterministic role reallocations, absolute fused-minus-audio
accuracy differences range from $-6.00\%$ to $+0.24\%$ (mean $-2.35\%$).

\Needspace{3\baselineskip}
This result does not support an encoder-general temporal-coverage effect under
these fixed checkpoints and does not isolate which encoder property explains
the difference.

Full-modality value does not establish predictable acquisition value. In the
eight-frame condition, the calibrated audio-to-video router requests video for
99.36\% of singleton evaluation clips. Its accuracy is 50.96\%, versus
45.10\% for audio alone. Subtracting the unrounded scores gives an absolute
accuracy difference of $+5.87\%$. The router-minus-matched-random absolute
accuracy difference at the same request rate is only $+0.04\%$, with
a 16-actor bootstrap interval of [$-0.30\%$, $+0.36\%$]. Starting from video yields the
same pattern: the policy acquires audio for every evaluation clip. On all 1,300
evaluation clips, set-valued full audio--video accuracy rises from 34.95\% to
48.46\% in the voice--face-disagreement stratum; this post-query stratum is
diagnostic only and never a router input.

For this frozen feature pair, the audiovisual accuracy difference is positive
but uncertain; its calibrated router shows no selective value beyond matched
acquisition. Under HuBERT--DeiT, the audio-to-video router requests video on
71.70\% of clips and has an absolute accuracy difference from matched random selection of $+0.42\%$
[$-1.02\%$, $+1.87\%$]; the reverse router acquires audio for every clip. Thus
the initial-role second pair also supplies no selective-acquisition result.

Table~\ref{tab:cremad-main} collects the frozen-pair comparisons. Each
full-fusion and matched-random difference has a 95\% actor interval spanning
zero, so these results do not establish a reliable full-fusion or
selective-acquisition gain.

\begin{table}[h!]
\caption{Pinned CREMA-D comparison on 1,244 singleton-vote clips from 16 held-out actors. Pair names list audio then video encoders; A+V uses both. Audio and A+V columns report accuracy, and request is the realized audio-to-video router rate. The two difference columns report A+V minus audio and router minus matched-random accuracy, respectively, computed from unrounded scores as absolute differences (\%). Brackets give 95\% actor bootstrap intervals; matched random uses the same request rate.}
\label{tab:cremad-main}
\vspace{0.4em}
\centering
\footnotesize
\setlength{\tabcolsep}{1.5pt}
\begin{tabularx}{\linewidth}{@{}>{\raggedright\arraybackslash}Xrrcrc@{}}
\toprule
\textbf{Frozen pair} & \textbf{Audio (\%) $\uparrow$} & \textbf{A+V (\%) $\uparrow$} &
\shortstack{\textbf{A+V $-$ audio}\\\textbf{(\%) $\uparrow$}} &
\textbf{Request (\%) $\downarrow$} &
\shortstack{\textbf{Router $-$ matched random}\\\textbf{absolute diff. (\%) $\uparrow$}} \\
\midrule
wav2vec 2.0--ViT & 45.10 & 50.96 & $+5.87$ [$-0.56$, $+11.77$] & 99.36 & $+0.04$ [$-0.30$, $+0.36$] \\
HuBERT--DeiT & 58.92 & 56.43 & $-2.49$ [$-8.40$, $+2.81$] & 71.70 & $+0.42$ [$-1.02$, $+1.87$] \\
\bottomrule
\end{tabularx}

\end{table}

An expanded $2\times2$ grid tests whether this boundary survives component
swaps. It joins wav2vec 2.0 or HuBERT audio features with ViT or DeiT visual
features and repeats the pre-label actor allocation six times. The
reallocations overlap, so their means and ranges are split-robustness
descriptions and not independent-population intervals. At calibration
budgets of 10, 25, 50, and 75\%, wav2vec 2.0--ViT has mean
absolute policy-minus-matched-random accuracy differences of $+0.54\%$,
$+1.24\%$, $+1.04\%$, and $+0.89\%$ at realized request rates of 5.46,
24.13, 44.23, and 65.50\%, respectively.

All six allocation means are
positive at 10\% and 25\%, but actor-bootstrap intervals span zero in five of
six allocations at 10\% and four of six at 25\%. At 50\% and 75\%, one of the
six allocation means is negative. The three
swapped pairs have no coherent sign pattern across budgets and reallocations.
Table~\ref{tab:cremad-comprehensive} makes the pair-specific means visible:
wav2vec 2.0--ViT has the largest descriptive margin at each calibration target,
but realized request rates differ across pairs.

\begin{table}[h!]
\caption{CREMA-D comprehensive frozen-feature grid. Values are means over six
overlapping pre-label actor reallocations. Column percentages are calibration
targets, not exact evaluation request rates. Full headroom is full audio--video
accuracy minus audio-only accuracy. $\Delta$ is policy accuracy minus expected
matched-random accuracy at the realized request rate; all values are absolute
accuracy differences (\%). The reallocation means are descriptive split robustness, not
independent-population estimates.}
\label{tab:cremad-comprehensive}
\vspace{0.4em}
\centering
\footnotesize
\setlength{\tabcolsep}{2.3pt}
\begin{tabularx}{0.82\linewidth}{@{}>{\raggedright\arraybackslash}p{2.4cm}rrrrr@{}}
\toprule
\textbf{Frozen pair} & \shortstack{\textbf{Full}\\\textbf{headroom (\%)} $\uparrow$} & \shortstack{\textbf{10\%}\\$\boldsymbol{\Delta}$ \textbf{(\%)} $\uparrow$} & \shortstack{\textbf{25\%}\\$\boldsymbol{\Delta}$ \textbf{(\%)} $\uparrow$} & \shortstack{\textbf{50\%}\\$\boldsymbol{\Delta}$ \textbf{(\%)} $\uparrow$} & \shortstack{\textbf{75\%}\\$\boldsymbol{\Delta}$ \textbf{(\%)} $\uparrow$} \\
\midrule
wav2vec 2.0 + ViT & $+6.28$ & $+0.54$ & $+1.24$ & $+1.04$ & $+0.89$ \\
wav2vec 2.0 + DeiT & $+3.09$ & $+0.01$ & $+0.16$ & $+0.34$ & $+0.05$ \\
HuBERT + ViT & $+0.89$ & $+0.14$ & $-0.13$ & $+0.25$ & $+0.68$ \\
HuBERT + DeiT & $-2.35$ & $+0.06$ & $+0.34$ & $+0.71$ & $+0.47$ \\
\bottomrule
\end{tabularx}

\end{table}

The other frozen pairs mix positive and negative margins across overlapping
reallocations. They do not establish an encoder-general selective-acquisition
result.

Table~\ref{tab:cremad-signs} reports how often each mean difference is positive
across the six overlapping actor reallocations. The wav2vec 2.0--ViT matched
margin is positive in all six allocations at 10\% and 25\%, but in five of six
at 50\% and 75\%; the other pairs show mixed signs across several targets.
These counts describe split sensitivity and are not independent replications.

\begin{table}[h!]
\caption{Sign consistency of the CREMA-D comprehensive grid. Each cell gives
the number of positive reallocation-level estimates out of six overlapping
actor allocations. Full headroom is full audio--video accuracy minus audio-only
accuracy; the four request columns give policy accuracy minus matched-random
accuracy at the realized request rate. Counts summarize split sensitivity and
are not confidence intervals or independent-population estimates.}
\label{tab:cremad-signs}
\centering
\footnotesize
\setlength{\tabcolsep}{3pt}
\begin{tabularx}{0.82\linewidth}{@{}>{\raggedright\arraybackslash}p{2.4cm}YYYYY@{}}
\toprule
\textbf{Frozen pair} & \textbf{Full headroom} & \textbf{10\%} &
\textbf{25\%} & \textbf{50\%} & \textbf{75\%} \\
\midrule
wav2vec 2.0 + ViT & 6/6 & 6/6 & 6/6 & 5/6 & 5/6 \\
wav2vec 2.0 + DeiT & 5/6 & 3/6 & 3/6 & 5/6 & 5/6 \\
HuBERT + ViT & 5/6 & 3/6 & 3/6 & 3/6 & 3/6 \\
HuBERT + DeiT & 1/6 & 3/6 & 4/6 & 5/6 & 5/6 \\
\bottomrule
\end{tabularx}
\end{table}

\FloatBarrier
\Needspace{9\baselineskip}
\subsection{AMI Audio--Video--Text Pilot}
\label{app:ami}
We evaluate frozen-feature dialogue-act classifiers using official AMI
signals and annotations \citep{carletta2005ami}. Ten meeting teams are
partitioned by role: five for task fitting, one for router fitting, one for
calibration, two for evaluation, and one for same-label donor controls. Unique
donor matching yields 244 of 616 candidate evaluation cases. Text features use
masked-mean pooling. We fit standardized one-hot ridge heads with $\alpha=1.0$
and measure observed encoder time for each modality. On these 244 donor-matchable cases,
the two-team macro accuracies are 60.62\% for audio, 44.60\% for audio--video,
76.18\% for text, and 35.77\% for audio--video--text. Adding modalities to
this fitted head therefore reduces accuracy; text-only accuracy does
not imply a usable text-request policy. The learned router uses only available audio--video features and head outputs,
is fit on router-fit teams, and selects its threshold on calibration teams
subject to a 50\% request cap.
It requests text on 38 of 244 evaluation records (15.57\%). The router-minus-static
absolute difference in accuracy is $-3.69\%$, averaged over the two evaluation
teams. At exactly 38 requests, the router-minus-random and router-minus-confidence
absolute differences in accuracy are $-1.69\%$ and $-4.96\%$, respectively. A
label-informed oracle at this same request count has an absolute difference in
accuracy of $+15.56\%$ versus static; it is an unattainable reference, not evidence that the learned router
can identify those records. Mean observed text extraction time is 4.28~ms
and audio--video extraction is 1404.26~ms on these records. These modality
measurements do not include complete application latency.

The macro values above average the two held-out meeting-team accuracies with
equal weight: $(\mathrm{accuracy}_{\text{ES2006}}+
\mathrm{accuracy}_{\text{ES2011}})/2$. Table~\ref{tab:ami-team-scores} gives
the observed class counts and all seven subset accuracies by team;
Table~\ref{tab:ami-router-scores} gives the selected router and its exact-count
controls by team. This exposes the two terms behind each macro value; a
pooled accuracy over all 244 records would weight the teams differently.

\begin{table}[h!]
\caption{AMI accuracy by evaluation meeting team. Class counts follow
the observed target-class order \texttt{elicit}, \texttt{minor},
\texttt{other}, \texttt{task}. Accuracy is the percentage correct within each
team for the indicated feature subset; A, V, and T denote audio, video, and
text. The two-team macro is the unweighted
mean of the ES2006 and ES2011 values; the breakdown is descriptive because only
two evaluation teams are available.}
\label{tab:ami-team-scores}
\centering
\footnotesize
\setlength{\tabcolsep}{2pt}
\begin{tabularx}{\linewidth}{@{}l r p{2.5cm}YYYYYYY@{}}
\toprule
\textbf{Team} & \textbf{Records} & \textbf{Class counts} &
\multicolumn{7}{c}{\textbf{Accuracy (\%) $\uparrow$}} \\
\cmidrule(lr){4-10}
& & & \textbf{A} & \textbf{V} & \textbf{T} & \textbf{A+V} &
\textbf{A+T} & \textbf{V+T} & \textbf{A+V+T} \\
\midrule
ES2006 & 123 & 9 / 23 / 11 / 80 & 65.04\% & 27.64\% & 82.11\% & 52.85\% & 64.23\% & 34.96\% & 21.95\% \\
ES2011 & 121 & 9 / 23 / 10 / 79 & 56.20\% & 19.83\% & 70.25\% & 36.36\% & 53.72\% & 37.19\% & 49.59\% \\
\bottomrule
\end{tabularx}
\end{table}

\begin{table}[h!]
\caption{AMI router accuracy by evaluation team for the available
audio--video baseline and text-request policies. Each percentage is computed
within the team's records. Each exact-count control uses 38 requests across
all 244 records; team allocations can differ. Macro differences described in
the text are unweighted means of the two team-level differences.}
\label{tab:ami-router-scores}
\centering
\footnotesize
\setlength{\tabcolsep}{3pt}
\begin{tabularx}{0.84\linewidth}{@{}l X r@{}}
\toprule
\textbf{Team} & \textbf{Policy} & \textbf{Accuracy (\%) $\uparrow$} \\
\midrule
ES2006 & Audio--video static & 52.85\% \\
 & Learned text router & 48.78\% \\
 & Matched random & 44.72\% \\
 & Matched confidence & 48.78\% \\
 & Label-informed oracle & 69.92\% \\
\midrule
ES2011 & Audio--video static & 36.36\% \\
 & Learned text router & 33.06\% \\
 & Matched random & 40.50\% \\
 & Matched confidence & 42.98\% \\
 & Label-informed oracle & 50.41\% \\
\bottomrule
\end{tabularx}
\end{table}

\FloatBarrier

This AMI study is exploratory because only two meeting teams contribute
evaluation cases. A common-rate comparison of the static, all-request, and
selective policies is infeasible because their request rates are fixed by their
definitions. We therefore report observed rates and exact-count matched
controls; these results do not establish a natural-source acquisition gain.

Across the descriptor, audiovisual, and meeting studies, full-modality
accuracy does not establish a useful pre-request ranking. The studies also
have different targets and independent units, so their accuracy differences
should not be pooled as one transfer estimate.

\raggedbottom
\FloatBarrier
\section{Matched Synthetic Controls}
\label{app:controls}
The five-seed control isolates the action identity of the fixed threshold
policy. Always-query-selective uses the same final-action rule but pays for
every request; query-majority keeps the request schedule and changes the
post-request action.
Table~\ref{tab:controls} shows that the active and always-query-selective
policies make identical decisions at different request rates, while
query-majority keeps the active request rate but has lower utility after
changing the post-request action.

\begin{table}[H]
\centering\footnotesize
\caption{Matched synthetic controls at $\sigma=0.22$, $\tau=0.35$. Targeted decision accuracy and utility are mean (sample standard deviation) over five paired seeds. All methods start with the same pair; all requests cost 0.10.}
\label{tab:controls}
\vspace{0.4em}
\setlength{\tabcolsep}{1.5pt}
\begin{tabularx}{0.76\linewidth}{@{}>{\raggedright\arraybackslash}p{2.3cm}YYYY@{}}
\toprule
\textbf{Policy} & \shortstack{\textbf{Targeted}\\\textbf{acc.} $\uparrow$} & \shortstack{\textbf{Utility}\\$\uparrow$} & \shortstack{\textbf{Query}\\$\downarrow$} & \textbf{Coverage}\\
\midrule
Pair-selective & \meanstd{0.594}{0.006} & \meanstd{0.110}{0.012} & $0.000$ & $0.366$\\
Query-majority & \meanstd{0.628}{0.005} & \meanstd{0.192}{0.010} & $0.634$ & $1.000$\\
Always-query-selective & \meanstd{0.790}{0.003} & \meanstd{0.447}{0.007} & $1.000$ & $0.594$\\
\rowcolor{green!8}\ours{} & \meanstd{0.790}{0.003} & \meanstd{0.484}{0.006} & $0.634$ & $0.594$\\
\bottomrule
\end{tabularx}
\end{table}
\FloatBarrier

The matched-action identity is direct evidence for request savings. It does
not imply that the initial pair predicts which third scores will change an
answer: under this terminal rule, no third score changes the action when the
threshold policy declines to request it.

\Needspace{9\baselineskip}
\section{Matched-budget Selector Value}
\label{app:selector}
The 83-seed comparison applies one terminal rule to train-only value,
pair-uncertainty, matched-random, and oracle request schedules at four budgets.
The no-query reference has utility $-0.0459$ (sample standard deviation
$0.0160$) at each budget. The final column reports the train-only selector's
paired mean utility difference from matched random, with a 95\% paired-seed
percentile interval from 10,000 resamples using seed 92309. Positive values
favor learned value selection. Pair uncertainty has higher utility than the
learned selector at all four budgets. The oracle uses held-out labels and the
unavailable score, so it is an upper bound at each exact budget and is not a deployable policy.
An exact budget forces requests even when their gain is below the request
cost; the oracle utility may therefore fall as the budget grows.
Table~\ref{tab:matched-selector} shows the budget dependence: the learned
selector exceeds matched random at 10\% and 25\%, falls below no-query utility
at 50\% and 63.7\%, and remains below pair uncertainty in utility at every
budget. The two high-budget comparisons reverse for non-ambiguous binary
accuracy: learned versus pair uncertainty is 0.4922 versus 0.4827 at 50\%
and 0.4443 versus 0.4296 at 63.7\%. This accuracy excludes ambiguity cases
and scores abstentions as errors. It does not encode request or abstention cost,
and targeted decision accuracy favors pair uncertainty at both budgets.
Among requested examples, mean pre-query-cost reward change from the third
score is $-0.0591$ versus $+0.1756$ at 50\% and $-0.1343$ versus $+0.0235$
at 63.7\% (learned versus pair uncertainty). This is the change between
the common terminal rule's full-score and pair-only actions before request
cost, averaged over requested examples within each of 83 seeds. Because
the schedules request different examples, it is a conditional diagnostic,
not a paired population effect or evidence that the third score is harmful
for all examples; repeated requests are not independent population units.

\begin{table}[H]
\centering
\footnotesize
\caption{Matched-budget selector comparison under one shared terminal rule
over 83 seeds. Cells show utility mean (sample standard deviation). The final
column gives the train-only value selector's paired mean utility difference
from matched random with a 95\% paired-seed percentile interval. The no-query
reference is $-0.0459$ ($0.0160$) at each budget.}
\label{tab:matched-selector}
\vspace{0.4em}
\setlength{\tabcolsep}{1.8pt}
\begin{tabularx}{0.98\linewidth}{@{}p{1.1cm}YYYY>{\centering\arraybackslash}p{4.0cm}@{}}
\toprule
\textbf{Query budget} &
\shortstack{\textbf{Train-only}\\\textbf{value} $\uparrow$} &
\shortstack{\textbf{Pair}\\\textbf{uncertainty} $\uparrow$} &
\shortstack{\textbf{Matched}\\\textbf{random} $\uparrow$} &
\shortstack{\textbf{Oracle}\\\textbf{upper bound} $\uparrow$} &
\shortstack{\textbf{Value--random}\\\textbf{$\Delta$ U (95\% CI)} $\uparrow$} \\
\midrule
10\% & \meanstd{-0.0321}{0.0166} & \meanstd{-0.0304}{0.0164} & \meanstd{-0.0570}{0.0168} & \meanstd{0.1479}{0.0159} & \shortstack{$+0.0249$\\$[0.0233,\,0.0264]$} \\
\addlinespace[0.9em]
25\% & \meanstd{-0.0418}{0.0161} & \meanstd{-0.0296}{0.0166} & \meanstd{-0.0721}{0.0164} & \meanstd{0.2455}{0.0151} & \shortstack{$+0.0303$\\$[0.0280,\,0.0325]$} \\
\addlinespace[0.9em]
50\% & \meanstd{-0.1254}{0.0175} & \meanstd{-0.0081}{0.0147} & \meanstd{-0.0994}{0.0164} & \meanstd{0.2205}{0.0151} & \shortstack{$-0.0260$\\$[-0.0292,\,-0.0228]$} \\
\addlinespace[0.9em]
63.7\% & \meanstd{-0.1952}{0.0167} & \meanstd{-0.0946}{0.0151} & \meanstd{-0.1148}{0.0158} & \meanstd{0.2068}{0.0151} & \shortstack{$-0.0804$\\$[-0.0832,\,-0.0776]$} \\
\bottomrule
\end{tabularx}

\end{table}
\FloatBarrier

The low-budget gain over matched random is limited to this shared terminal
rule and synthetic population. At larger budgets the learned ranking requests
examples with negative mean reward change, while pair uncertainty retains
higher utility despite the learned ranking's small binary-accuracy advantage.
Accuracy and costed utility answer different questions here.

\Needspace{5\baselineskip}
\section{Cost-Sensitivity Detail}
\label{app:cost}
A separate $3\times3$ grid varies score noise
$\sigma\in\{0.11,0.22,0.44\}$ and threshold
$\tau\in\{0.175,0.35,0.7\}$ without test-set threshold selection.
Active utility spans $-0.058$ to $0.497$.

At campaign noise, raising $\tau$ from 0.35 to 0.7 changes utility
from 0.484 to $-0.039$; at $\tau=0.35$, raising noise to 0.44 gives
0.337. These shifts reflect
score magnitudes and the weak-score ambiguity marker, not robustness to
arbitrary rescaling.

\begin{figure}[H]
  \centering
  \includegraphics[width=0.58\linewidth]{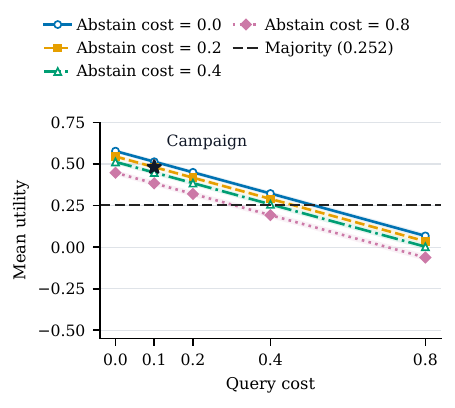}
  \caption{Cost sensitivity on the partially held-out split. Curves show the
  active policy's mean utility as query cost varies; shaded regions are one
  standard deviation across seeds. The dashed line is the majority baseline,
  whose query and abstention rates are zero. The star marks the campaign costs.}
  \label{fig:cost}
\end{figure}

Figure~\ref{fig:cost} asks how much of the threshold rule's synthetic utility
survives higher request prices. Its utility decreases as $\lambda_q$ rises,
while majority stays constant because it neither queries nor abstains. This
is a re-scoring of fixed actions, not a policy refit.

Table~\ref{tab:cost} gives the companion numerical comparison. At the campaign
costs $(\lambda_q,\lambda_a)=(0.10,0.20)$, the threshold rule has
$0.481\pm0.014$ mean utility versus $0.252\pm0.016$ for majority. At
$(0.40,0.80)$, its utility is $0.192\pm0.018$, below majority's unchanged
$0.252\pm0.016$. The cost pair changes the preferred policy under this
generator; the analysis does not estimate real acquisition latency or cost.

\begin{table}[H]
\caption{Cost sensitivity in the 83-seed scalar study. Policy utilities are
mean (sample standard deviation); the utility-difference column is threshold utility
minus majority utility. Majority is unchanged because it never requests or
abstains. Costs are query $\lambda_q$ and non-ambiguous abstention
$\lambda_a$.}
\label{tab:cost}
\vspace{0.4em}
\centering
\footnotesize
\setlength{\tabcolsep}{1.7pt}
\begin{tabularx}{0.78\linewidth}{@{}>{\raggedright\arraybackslash}p{1.9cm}YYY@{}}
\toprule
\shortstack{\textbf{Costs}\\$(\lambda_q,\lambda_a)$} & \shortstack{\textbf{Threshold}\\\textbf{utility} $\uparrow$} & \shortstack{\textbf{Majority}\\\textbf{utility} $\uparrow$} & \shortstack{\textbf{Utility}\\\textbf{difference} $\uparrow$} \\
\midrule
$(0.00, 0.00)$ & \meanstd{0.578}{0.012} & \meanstd{0.252}{0.016} & $0.326$ \\
$(0.10, 0.20)$ & \meanstd{0.481}{0.014} & \meanstd{0.252}{0.016} & $0.230$ \\
$(0.20, 0.40)$ & \meanstd{0.385}{0.015} & \meanstd{0.252}{0.016} & $0.133$ \\
$(0.40, 0.80)$ & \meanstd{0.192}{0.018} & \meanstd{0.252}{0.016} & $-0.059$ \\
\bottomrule
\end{tabularx}

\end{table}

Figure~\ref{fig:cost} and Table~\ref{tab:cost} locate a change in policy
ordering under fixed actions. They do not identify application-specific
acquisition prices, and no threshold is refitted for the alternative costs.

\Needspace{5\baselineskip}
\section{Extended Limitations}
\label{app:limitations}
The study isolates a protocol, and that isolation determines what the results
mean. Scores are not waveforms, frames, or language-model representations. The
active policy receives a third score after a request without modeling
acquisition latency, missingness, privacy, or temporal alignment.

The CREMA-D
routing comparison matches request rate only; it does not measure decoding
delay, monetary cost, or a deployment availability process. The synthetic
generator also uses a fixed threshold and a small set of mechanisms. Invert and
swap share the same score-level transformation.

The reported mechanism-label summaries retain both labels and therefore count
that transformation twice.
Future studies could test distinct transformations.

The temporal extension is not a natural-media evaluation. Its zero-mean marker
removes information from time means while leaving a fixed waveform in full
traces. The equal-information follow-up shows that a train-only template and a
ridge classifier match the temporal networks and that a shared time permutation
preserves the separation. The original factorial therefore does not establish a
temporal-model advantage or transfer to raw audio, video, or text.

\paragraph{Permutation diagnostics.}\label{app:temporal-permutations}
Under independent channel permutations, partially held-out-split targeted
decision accuracies are $0.768\pm0.009$, $0.619\pm0.009$, $0.428\pm0.015$, and
$0.428\pm0.008$ for temporal fusion, temporal gated, full-trace ridge, and the
train-only template, respectively. Per-example shifts give $0.781\pm0.009$,
$0.720\pm0.016$, $0.430\pm0.018$, and $0.424\pm0.007$ in the same order. The
shared train--evaluation permutation leaves all four methods within their
central-setting range because it preserves one fixed positional template.
The hard identity setting uses weaker signal and stronger noise but yields
similar utility across the methods (Table~\ref{tab:temporal}); it does not
establish a general difficulty gradient.

The learned baselines use only observed scores as model inputs and are trained
on the same synthetic family. Their seen performance is therefore a sanity
check, not a comparison with a state-of-the-art multimodal encoder. In
particular, the active rule has access to the third coordinate after a request,
whereas fixed baselines always receive all three coordinates at the initial
decision. The CREMA-D
frozen-backbone replication also reverses the positive wav2vec 2.0--ViT fusion
difference, so the natural-media result is not encoder-independent. A fair
real-media comparison would need to define modality availability, request
latency, preprocessing cost, and a preregistered encoder set for every method.

The designated ambiguity mechanism encodes a particular safety preference:
abstention is rewarded for these examples, but not for the equally
label-nonidentifiable mixed examples. Changing $\lambda_q$ or $\lambda_a$
changes the preferred operating point. We therefore report the underlying
action rates and selective accuracy alongside utility, and we treat cost
sensitivity as a required stress test before interpreting a policy as
preferable.

The descriptor study uses released features but does not resolve the score
abstraction. Validation-only calibration largely closes the fixed
threshold's accuracy gap, but does not improve utility over the matched
majority control. The Perception Test pilot does not establish a usable conflict benchmark: its question pool fails the source-grounding criteria, and its 20-component decision partition includes only 12 unseen components, below the prespecified minimum of 15. These results do not show that other Perception Test questions or multimodal models lack source-specific information. A new raw study needs a separately frozen dataset,
model, and task contract. The study uses previously released benchmark data and
derived descriptors; it recruits no participants, gathers no new annotations,
and runs no human evaluation. It inherits the representational and consent
limitations of those releases. Human evaluation is future work, subject to
explicit approval; the present results do not address perceptual quality,
accessibility, clinical use, or deployment readiness.

\end{document}